\documentclass{article}

\usepackage{arxiv}

\usepackage[utf8]{inputenc} 
\usepackage[T1]{fontenc}    
\usepackage{hyperref}       
\usepackage{url}            
\usepackage{booktabs}       
\usepackage{amsfonts}       
\usepackage{nicefrac}       
\usepackage{microtype}      
\usepackage{lipsum}
\usepackage{graphicx}
\graphicspath{ {./images/} }

\title{An Aspect Extraction Framework using Different Embedding Types, Learning Models, and Dependency Structure}

\author{
 Ali Erkan \\
  Computer Engineering Department\\
  Boğaziçi University\\
  İstanbul, Turkey \\
  \texttt{alierkan@gmail.com} \\
   \And
 Tunga Güngör\\
  Computer Engineering Department\\
  Boğaziçi University\\
  İstanbul, Turkey \\
  \texttt{gungort@boun.edu.tr} \\
}

\usepackage{pgfplots}
\pgfplotsset{compat=newest}
\pgfplotsset{
	tick label style = {font = \normalsize},
	label style = {font = \normalsize},
	legend style = {font = \normalsize},
	every axis plot/.append style = {font = \normalsize},
	xticklabel style = {font=\normalsize}
}
\usetikzlibrary{matrix,chains,positioning,decorations.pathreplacing,arrows}

\graphicspath{ {./images/} }

\usepackage{listings}
\graphicspath{ {./images/} }
\usepackage{array, multirow, multicol}
\newcolumntype{C}[1]{>{\centering\arraybackslash}p{#1}}
\newcolumntype{L}[1]{>{\raggedright\arraybackslash}p{#1}}
\usepackage{booktabs}

\newcommand*{\splitl}[1]{%
	\begingroup
	\renewcommand*{\arraystretch}{1.1}%
	\begin{tabular}[c]{@{}l@{}}#1\end{tabular}%
	\endgroup
}

\begin{document}
\maketitle
\begin{abstract}
Aspect-based sentiment analysis has gained significant attention in recent years due to its ability to provide fine-grained insights for sentiment expressions related to specific features of entities. An important component of aspect-based sentiment analysis is aspect extraction, which involves identifying and extracting aspect terms from text. Effective aspect extraction serves as the foundation for accurate sentiment analysis at the aspect level. In this paper, we propose aspect extraction models that use different types of embeddings for words and part-of-speech tags and that combine several learning models. We also propose tree positional encoding that is based on dependency parsing output to capture better the aspect positions in sentences. In addition, a new aspect extraction dataset is built for Turkish by machine translating an English dataset in a controlled setting. The experiments conducted on two Turkish datasets showed that the proposed models mostly outperform the studies that use the same datasets, and incorporating tree positional encoding increases the performance of the models.
\end{abstract}

\section*{Introduction}
Aspect-based sentiment analysis (ABSA) is a subfield of natural language processing (NLP) that aims to identify the sentiment expressed towards specific aspects or features of a product, service, or entity. The task involves analyzing the text to identify the various aspects or attributes being discussed and determining the sentiment expressed towards each aspect. ABSA is an essential area of research as it has many practical applications in industries such as e-commerce, social media monitoring, and customer service. ABSA is a rapidly evolving field with new techniques and approaches being proposed regularly. As businesses continue to place increasing importance on customer feedback and sentiment analysis, ABSA is likely to play an increasingly important role in the e-commerce and customer service industries.

ABSA is a featured subfield of sentiment analysis that has received significant attention in recent years \cite{Zhang2022, chauhan2023}. Traditional sentiment analysis focuses on determining the overall polarity of a piece of text, classifying it as positive, negative, or neutral. However, in many real-world scenarios, this global sentiment classification does not provide a comprehensive view of the different kinds of opinions expressed in the text.

ABSA addresses this limitation by offering a more granular approach to sentiment analysis. Instead of treating the entire text as a single unit, ABSA aims to analyze sentiments concerning specific aspects or features of the discussed entities. Aspects can refer to particular attributes, components, or elements associated with entities such as products, services, or experiences. In a product review, for instance, ABSA enables us to observe the sentiments expressed about individual features of the product, such as its design, performance, battery life, and customer service. This fine-grained analysis provides companies and researchers with more detailed information and reflects the nuanced nature of human opinions. Table \ref{tab:sample_aspects} shows some examples of customer restaurant reviews and the aspects of the reviews.

\begin{table}[!ht]
	\centering
		\small
		\caption{Examples of Turkish customer restaurant reviews and aspects in the reviews.}
		\label{tab:sample_aspects}
		\begin{tabular}{|p{330pt}|p{90pt}|}
			\hline
			\textbf{Review} Turkish (English) & \textbf{Aspects}\\
			\hline 
			\begin{tabular}{l} gördüğüm en güzel dekorasyonlu simit sarayı diyebilirim. \\ (I can say that it is Simit Palace with the most beautiful decoration I have ever seen.)\end{tabular} & 
			\begin{tabular}{l} dekorasyonlu \\ (decoration) \end{tabular}
			\\ \hline
			\begin{tabular}{l} lokumu tavsiye ederim. \\ (I recommend Turkish delight.) \end{tabular} & 
			\begin{tabular}{l} lokumu \\ (Turkish delight) \end{tabular}
			\\ \hline
			\begin{tabular}{l} kahvaltı tabağının 12.5 tl den 18.5 ye çıkması gerçekten çok komik. \\ (It's really funny that the breakfast plate goes from 12.5 TL to 18.5 TL.) \end{tabular} & 
			\begin{tabular}{l} kahvaltı tabağının \\ (breakfast plate) \end{tabular}
			\\ \hline
			\begin{tabular}{l} tüm hardallı balığı denedim ardından elmalı yeşil çay ikisi de enfesti. \\ (I tried all the fish with mustard and then the green tea with apple, both were delicious.) \end{tabular} & 
			\begin{tabular}{l} hardallı balığı \\ (fish with mustard) \\ elmalı yeşil çay \\ (green tea with apple)\end{tabular}
			\\ \hline
		\end{tabular}
\end{table}

The fundamental challenge in ABSA lies in accurate aspect extraction, which aims to identify and extract the aspect terms present in the text. This step forms the basis for subsequent sentiment analysis, as sentiments are linked to specific aspects. The extraction of aspects involves identifying keywords, phrases, or syntactic patterns that refer to the aspects under consideration.

In this paper, we focus on aspect extraction, which is a pivotal task in understanding the fine-grained nuances of sentiments. Using the synergy of cutting-edge techniques, we propose a comprehensive model that integrates BERT (Bidirectional Encoder Representations from Transformers) \cite{bert2018}, BiLSTM (Bidirectional Long Short-Term Memory) \cite{lstm97}, CRF (Conditional Random Fields) \cite{crf2001}, tree positional encoding (TPE), and different types of embeddings for enhanced aspect extraction. By integrating different types of models, we aim to handle the different requirements in aspect extraction.

BERT, renowned for its contextualized word representations, allows the model to capture intricate semantic relationships within the text. BiLSTM, with its ability to model long-term dependencies, addresses the hierarchical and sequential nature of aspects, which is essential for accurate extraction. Incorporating CRF as a final layer augments the model's performance by considering the sequential dependencies between the terms and thus imposing constraints between adjacent aspect predictions. Furthermore, exploring pre-trained word embeddings and tree positional encoding in conjunction with the neural network architecture enhances the model's sensitivity to linguistic subtleties, contributing to a more nuanced understanding of aspects.

The motivation for this research stems from the recognition that existing ABSA models, while successful to varying degrees, often face challenges in handling the complexity of real-world textual data. By combining state-of-the-art techniques, our approach seeks to push the boundaries of aspect extraction precision and contribute to a more nuanced understanding of sentiments in textual information.

The main contributions in this paper are as follows: \begin{itemize}
	\item We develop a novel tree positional encoding model that is based on dependency parse trees.
	\item We combine different types of learning algorithms (BERT, BiLSTM, and CRF) that contribute to different aspects of the task. As input to the models, we use word and part-of-speech (POS) embeddings and tree positional encodings.
	\item We obtain state-of-the-art results for extracting aspects from Turkish review datasets.
	\item We release and make publicly available\footnote{https://github.com/alierkan/Turkish-ABSA} a new aspect-based sentiment analysis dataset for Turkish. It can serve as a benchmark dataset for Turkish ABSA studies. We also share our Python codes. 
\end{itemize}

The rest of the paper is organized as follows: Section \ref{sec:literature} briefly reviews the literature on aspect extraction. Section \ref{sec:datasets} presents the datasets used in this work and the processes involved in translating the English reviews dataset into Turkish. Section \ref{sec:model} explains in detail the proposed models, the input schema used in the models, and the evaluation metrics. Section \ref{sec:experiments} describes the experimental settings and presents the results. Finally, Section \ref{sec:conclusion} concludes the paper.

\section{Literature Review}
\label{sec:literature}

In recent years, there has been a significant amount of research in aspect-based sentiment analysis and several different approaches and methods have been proposed. In this section, we review some of the key works and developments in this domain.

Early ABSA studies employed rule-based methods for aspect extraction. Various techniques and resources, such as pattern matching, syntactic parsing, and domain-specific lexicons, were used. Hu and Liu \cite{hu2004} proposed a method for identifying the sentiment of product reviews based on the presence of specific keywords and phrases. They introduced a rule-based approach using POS tagging and dependency parsing to extract aspects from reviews. This approach, known as the lexicon-based approach, has since been widely used in ABSA research. Qiu et al. \cite{Qiu2011} proposed a method by incorporating the WordNet lexical database into the model.

Following these works, many researchers have proposed more advanced techniques for ABSA. Some of the works focused on using machine learning algorithms to identify aspects and their corresponding sentiments in the text. As an example in this direction, Jihan et al. \cite{Jihan2017} proposed an SVM-based approach using the mean embedding feature to extract all related aspect categories. After spell correction, they obtained lemmatized words and then used mean vectors of Word2Vec embeddings of the lemmas as the features in the SVM model. In addition to classical machine learning approaches, deep learning models have also begun to be used for aspect extraction. Different models based on recurrent neural networks (RNNs) \cite{rnn85} and convolutional neural networks (CNNs) \cite{cnn98} have been proposed. Poria et al. \cite{poria2016} introduced a CNN-based model for aspect extraction using semantic and sentiment information. They used Word2Vec embeddings of words and part-of-speech tags as features and also combined the CNN model with a rule-based algorithm based on linguistic patterns. Wang et al. \cite{wang2016} proposed an LSTM-based approach that uses attention mechanisms to identify important aspects and their associated sentiments. The input of the LSTM model was word embeddings and the output of the LSTM model fed the attention mechanism. The attention mechanism was used to take into account different parts of the sentence for different aspects. In a recent survey, Chauhan et al. \cite{chauhan2023} summarized aspect-based sentiment analysis studies that use deep learning models.

Attention mechanisms and transformer architectures have made a significant leap in aspect extraction. As a pioneering model, BERT introduced contextualized embeddings that improved aspect extraction by considering surrounding words and their impact on aspect terms. The transfer learning paradigm was harnessed by fine-tuning pre-trained models like BERT. In a work that followed this approach, BERT model was fine-tuned for aspect extraction, demonstrating the benefits of transfer learning in capturing domain-specific nuances \cite{finebert2019}.

Hybrid approaches gained traction in aspect-based studies, combining rule-based, supervised, and deep learning techniques. Chauhan et al. \cite{chauhan2020} presented a hybrid and domain-based model that integrates rule-based methods and neural networks, achieving improved aspect extraction accuracy. The single word and multi-word aspects are first extracted using a set of linguistic rules. The aspects are pruned based on the frequency and semantic similarity criteria to eliminate the terms that are not relevant to the domain. Then a BiLSTM network is trained using the extracted aspect terms.

Another vital aspect of ABSA is its multilingual and cross-lingual nature. As businesses operate increasingly globally, there is a growing need for ABSA techniques that can handle multiple languages. To this end, researchers have proposed various methods such as cross-lingual transfer learning \cite{li2022} and multilingual neural networks \cite{insig2016} to improve ABSA performance in different languages. Li et al. \cite{li2022} developed a model that focused on increasing attention to domain-invariant features while reducing attention to domain-specific features. They aimed to determine how closely related tokens are regarding the type of inference. They proposed a kernelized model with a kernelized attention mechanism that uses a shortest-path graph kernel to constrain the attention scope of variables. They used a metagrammar representation for input features with BERT embedding and they assembled kernelized and non-kernelized models. Ruder et al. \cite{insig2016} first collected 10,000 most frequent words as the vocabulary for each language. Then they used GloVe embedding vectors for English and random embedding vectors for other languages. They trained a CNN model to extract aspects.

To evaluate aspect extraction methods, benchmark datasets are widely used. SemEval datasets \cite{semeval2014, semeval2016} provide a standardized basis for performance comparison. Metrics such as precision, recall, F1 score, and accuracy are commonly used for evaluation.

As a summary of the literature review, we observe that aspect extraction is a crucial first step in aspect-based sentiment analysis, affecting the accuracy and granularity of sentiment analysis. The evolution from rule-based methods to deep learning methods, mainly influenced by pre-trained language models like BERT, has transformed aspect extraction. Using hybrid techniques and integrating contextual and linguistic information into the models are promising directions to improve the accuracy of aspect extraction.

\section{Datasets}
\label{sec:datasets}
In this work, we use two datasets for aspect extraction to analyze the effects of different model architectures and embedding types on Turkish reviews. Both datasets are SemEval shared task datasets that are publicly available\footnote{https://alt.qcri.org/semeval2016/task5/index.php?id=data-and-tools} and are commonly used as benchmark datasets in semantically related NLP tasks (e.g. \cite{ettaleba2022, kanwal203, alharbi2024}). The first is the SemEval 2016 Task 5 Restaurant Turkish Reviews dataset \cite{semeval2016}. As a second dataset, we machine translate the Semeval 2016 Task 5 Restaurant English Reviews dataset \cite{semeval2016} into Turkish. To translate the dataset, we used Googletrans\footnote{https://pypi.org/project/googletrans}, which is a Python library that uses the Google Translate service\footnote{https://translate.google.com/}. Table \ref{tab:datasets} shows the number of reviews in the training and test splits of the two datasets.

\begin{table}[!ht]
	\centering
	\caption{Number of reviews in the datasets.}
	\label{tab:datasets}
	\setlength{\tabcolsep}{3pt}
	\begin{tabular}{|p{155pt}|p{40pt}|p{25pt}|p{25pt}|}
		\hline
		\textbf{Dataset} & \textbf{Training} & \textbf{Test} & \textbf{Total}\\
		\hline
		\begin{tabular}{l} SemEval 2016 English-Translated \\ Rest. Reviews \cite{semeval2016}
		\end{tabular} &
		\begin{tabular}{c} 2000 \end{tabular} & 
		\begin{tabular}{c} 676 \end{tabular} &
		\begin{tabular}{c} 2676 \end{tabular}
		\\ \hline
		\begin{tabular}{l} SemEval 2016 Turkish \\ Restaurant Reviews \cite{semeval2016}
		\end{tabular} &
		\begin{tabular}{c} 1104 \end{tabular} & 
		\begin{tabular}{c} 144 \end{tabular} &
		\begin{tabular}{c} 1248 \end{tabular}
		\\ \hline
		\begin{tabular}{l} \textbf{Total}
		\end{tabular} &
		\begin{tabular}{c} 3104 \end{tabular} & 
		\begin{tabular}{c} 820 \end{tabular} &
		\begin{tabular}{c} 3924 \end{tabular}
		\\ \hline
	\end{tabular}
\end{table}

The reasons of compiling a new dataset are threefold. First, the size of the Turkish restaurant reviews dataset is low compared to all the other restaurant reviews datasets in the shared task and we aim at increasing the number of reviews. Second, related to the first goal, we can increase the performance of the models by increasing the size of the training set. The third and most important motivation is that we contribute a new large-scale benchmark dataset for aspect-based sentiment analysis research in Turkish.

The SemEval Restaurant Reviews datasets are stored in XML format. Listing \ref{xml:sample} shows an example review from the English Restaurant Reviews dataset. A review is formed of a number of sentences. Each sentence includes zero or more aspect/sentiment information which is composed of an aspect ("target"), category of the aspect ("category"), sentiment for the aspect ("polarity"), and the position of the aspect within the sentence ("from"-"to"). 

\lstset{
	language=xml,
	tabsize=2,
	caption=An example review in XML format,
	label={xml:sample},
	frame=shadowbox,
	rulesepcolor=\color{gray},
	frame=none,
	keywordstyle=\color{black}\bf,
	commentstyle=\color{black},
	stringstyle=\color{black},
	numbers=none,
	breaklines=true,
	showstringspaces=false,
	basicstyle=\small,
	emph={Reviews,Review,sentences,sentence,text,Opinions,Opinion},emphstyle={\color{black}}}
\begin{lstlisting}
<?xml version="1.0" encoding="UTF-8" standalone="yes"?>
<Reviews>
    <Review rid="1090587">
        <sentences>
            <sentence id="1090587:2">
                <text>Add to that great service and great food at a reasonable price and you have yourself the beginning of a great evening.</text>
                <Opinions>
                    <Opinion target="service" category="SERVICE#GENERAL" polarity="positive" from="18" to="25"/>
                    <Opinion target="food" category="FOOD#QUALITY" polarity="positive" from="36" to="40"/>
                    <Opinion target="food" category="FOOD#PRICES" polarity="positive" from="36" to="40"/>
                </Opinions>
            </sentence>
            <sentence id="1090587:3">
                <text>The lava cake dessert was incredible and I recommend it.</text>
                <Opinions>
                    <Opinion target="lava cake dessert" category="FOOD#QUALITY" polarity="positive" from="4" to="21"/>
                </Opinions>
            </sentence>
        </sentences>
    </Review>
</Reviews>
\end{lstlisting}

To obtain a high-quality translated dataset, we performed several post-processing operations on the translated outputs. We list below some of the manual processing operations in order to guide future studies in translating datasets between different languages.

\begin{itemize}
	\item The machine translator system may translate the same source word in different reviews into different target words by taking into account the contexts of the words. However, it is important in aspect extraction to have the same concept (aspect) written in the same word form independent of the context. To solve this issue, we normalize all the usages of the same aspect in the reviews by choosing a canonical form of the word and replacing all occurrences with this form. For instance, the word "food" is sometimes translated as "yemek" and sometimes as "gıda". Food is an important aspect in the restaurant domain and should be written in a standard form in the reviews to detect it and the related sentiment. Therefore, in translated reviews, we replace all the "gıda" words in the translations of "food" with the "yemek" words.
	\item The translation may translate a word differently depending on whether the word appears within the review sentence or appears by itself as an aspect. For such cases where the translated aspect terms do not occur in the translated review sentence, we change the aspect terms by the corresponding terms in the sentence.
	\item English and Turkish belong to different language families and the word orders in the constituents differ greatly. This implies that the positions of the aspect terms in the source and target sentences will be different after the translation of a sentence. For all the sentences in the dataset, we relabel the aspect positions in the translated dataset by checking the review sentences. Table \ref{tab:samples} lists some examples showing how the positions of the aspect terms differ in the two languages.
\end{itemize}

\begin{table}[!ht]
		\centering
		\caption{Examples of English reviews and translated Turkish reviews with aspect positions updated.}
		\label{tab:samples}
		\setlength{\tabcolsep}{3pt}
		\begin{tabular}{|p{50pt}|p{200pt}|p{100pt}|p{60pt}|}
			\hline
			\begin{tabular}{l} \end{tabular} & 
			\begin{tabular}{l} \textbf{Review}  \end{tabular} & 
			\begin{tabular}{c} \textbf{Aspect}  \end{tabular} & 
			\begin{tabular}{l} \textbf{Aspect Pos.} \\ \textbf{in Review}  \end{tabular} \\
			\hline 
			\begin{tabular}{l} English \end{tabular} &
			\begin{tabular}{l} Judging from previous posts this used to be \\a good \textbf{place}, but not any longer.  \end{tabular} & 
			\begin{tabular}{c} place \end{tabular} &
			\begin{tabular}{c} 11 \end{tabular}
			\\ \hline
			\begin{tabular}{l} \textit{Translated} \end{tabular} &
			\begin{tabular}{l} \textit{Önceki gönderilerden yola çıkarak bu iyi bir} \\ \textit{\textbf{yerdi},} \textit{ama artık değil.} \end{tabular} & 
			\begin{tabular}{c} yerdi \end{tabular} &
			\begin{tabular}{c} 8 \end{tabular}
			\\ \hline  
			\begin{tabular}{l} English \end{tabular} &
			\begin{tabular}{l} I had the \textbf{duck breast special} on my last \\visit and it was incredible. \end{tabular} & 
			\begin{tabular}{c} duck breast special \end{tabular} &
			\begin{tabular}{c} 4,5,6 \end{tabular}
			\\ \hline
			\begin{tabular}{l} \textit{Translated} \end{tabular} &
			\begin{tabular}{l} \it Son ziyaretimde ördek göğsü özel \textit{vardı} \\ \textit{ve inanılmazdı.} \end{tabular} &
			\begin{tabular}{c} ördek göğsü özel \end{tabular} &
			\begin{tabular}{c} 3,4,5 \end{tabular}
			\\ \hline 
			\begin{tabular}{l} English \end{tabular} &
			\begin{tabular}{l} However, it's the \textbf{service} that leaves a \\ bad taste in my mouth. \end{tabular} & 
			\begin{tabular}{c} service \end{tabular} &
			\begin{tabular}{c} 4 \end{tabular}
			\\ \hline
			\begin{tabular}{l} \textit{Translated} \end{tabular} &
			\begin{tabular}{l} \textit{Ancak, ağzımda kötü bir tat bırakan} \\\textit{\textbf{hizmettir}.} \end{tabular} & 
			\begin{tabular}{c} hizmettir \end{tabular} &
			\begin{tabular}{c} 7 \end{tabular}
			\\ \hline
		\end{tabular}
\end{table}

\section{Aspect Extraction Models}
\label{sec:model}
\begin{figure}[th!]
	\centering
	\caption{\bf System architecture with BERT, BiLSTM, and CRF models}
	\includegraphics[width=\textwidth]{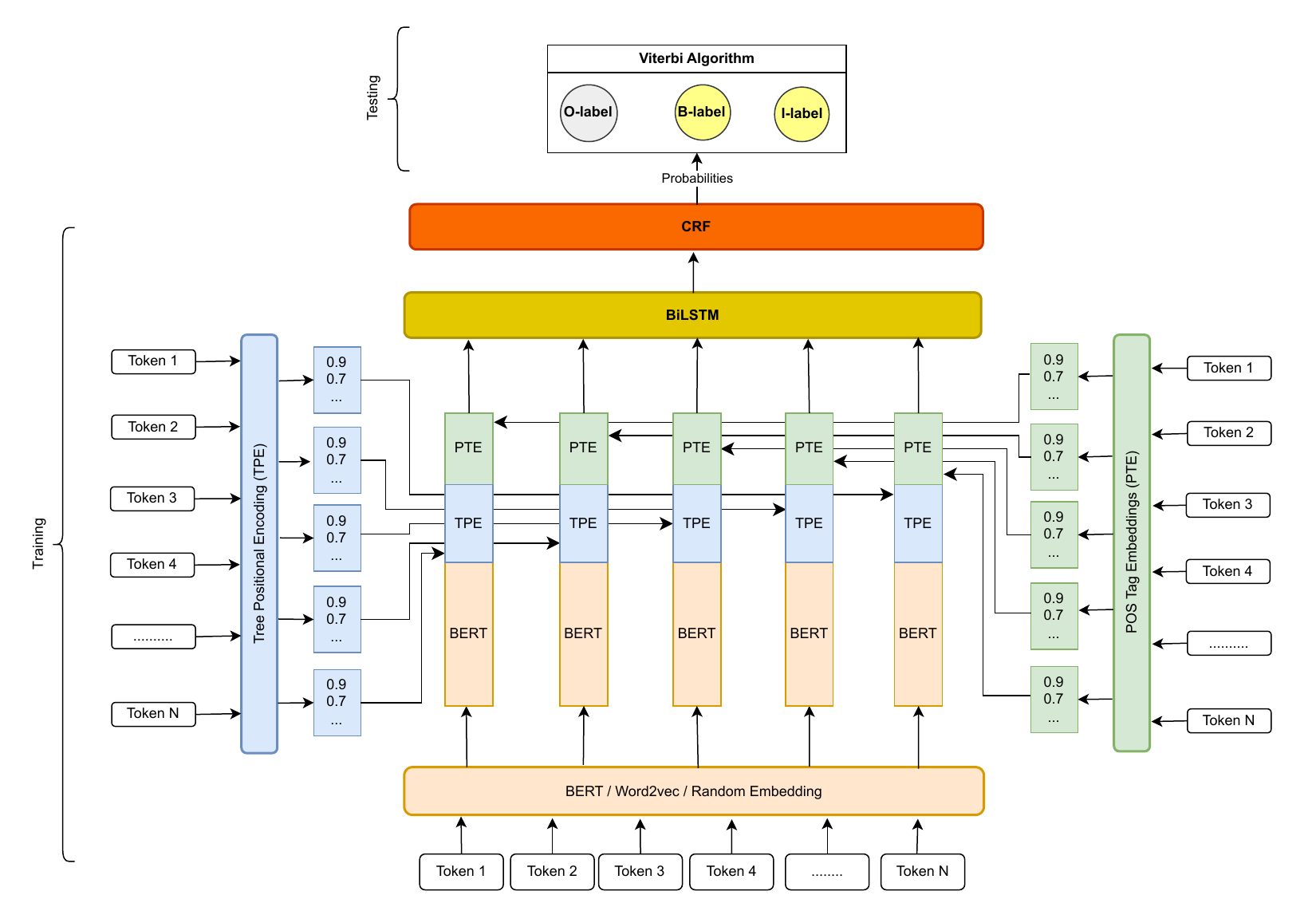}
	\label{fig:bert}
\end{figure}

In this section, we explain the aspect extraction models used in this work. We employ different model combinations with different types of embeddings. The general architecture is shown in Fig. \ref{fig:bert}. The models take as input information about the tokens in the review in the form of three types of embedding, which are word embeddings, POS tag embeddings, and tree positional encodings. The details of the models and the embedding schemes are explained in the following subsections.

\subsection{Word Embeddings}

We build two different models depending on how the word embeddings are fed into the model. In the first model, we use randomly initialized embeddings or domain-specific Word2Vec embeddings as word embeddings for the words in the review. We train the domain-specific Word2Vec embeddings using the two Turkish review datasets explained in Section \ref{sec:datasets}. We build a word embedding repository from these pretrained Word2Vec embeddings which are then used in the aspect extraction models. In both cases (random embeddings and Word2Vec embeddings), the word vectors are initialized with these embeddings and then updated during the training phase of the aspect extraction models. In this way, the word vectors are dynamically adjusted with respect to the aspect extraction task. We use 300-dimensional word vectors. In the second model, we use a Turkish BERT model ("bert-base-turkish-uncased"\footnote{https://huggingface.co/dbmdz/bert-base-turkish-uncased}) to produce the word embeddings. Instead of using random or pretrained embeddings as input to the model, we feed the words in the review as input to the BERT model and use the token outputs as word embeddings. As in the first model, the BERT embeddings are updated during the training phase of the aspect extraction models. The BERT base models have 756-dimensional vectors.

\subsection{POS Tag Embeddings}
\label{sec:POS}

Part-of-speech tags of the words in the reviews are obtained using the BOUN dependency parser\footnote{https://github.com/BOUN-TABILab-TULAP/Boun-Parser} \cite{Ozates2022}. The parser had been trained on three Turkish dependency treebanks in the Universal Dependencies (UD) framework\footnote{https://universaldependencies.org/}. It uses the POS tagset used in these treebanks, which is formed of 17 POS tags. As in the case of word embeddings, we use randomly initialized or domain-specific Word2Vec embeddings for the POS tag vectors of the words. The domain-specific POS tag embeddings are trained in the same manner as word embeddings. We represent the reviews in the datasets in terms of POS tags of the words, combine the reviews in the two datasets, and then train the Word2Vec embeddings using the Word2Vec algorithm. Random or Word2Vec embeddings are used to initialize the POS tag vectors in the models. The POS tag vectors are then dynamically updated during the training of the aspect extraction models. We use 100-dimensional POS tag vectors.

\subsection{Tree Positional Encodings}

In the original transformer model proposed by Vaswani et al. \cite{vaswani2017}, positional encodings are used to capture the position information of words in a sequence. For each word in the sequence, the word embedding vector and the positional embedding vector are added, which is then used as input to the attention layers. The dimension of the positional encoding is set equal to the dimension of the word vector. A static approach is used in forming the positional encoding that gives a unique encoding for each position in the sequence. The following equations are used in forming the positional encoding of position \textit{pos} ($PE_{pos}$) in a sequence:

\begin{equation}
	\label{eq:pe1}
	PE_{pos,2i} = \sin{\left(\frac{pos}{M^\frac{2i}{d}{}}\right)}
\end{equation}
\begin{equation}
	\label{eq:pe2}
	PE_{pos,2i+1} = \cos{\left(\frac{pos}{M^\frac{2i}{d}{}}\right)}
\end{equation}

\noindent where \textit{d} is the dimension of the position vector, $0 \leq i < \frac{d}{2}$ is used to index the elements of the vector, and \textit{M} is a user-defined constant which was set to 10,000 in the original architecture. As can be seen, the even-numbered elements of the vector are computed using the sine function, and the odd-numbered elements using the cosine function.

During our preliminary work in aspect extraction, we observed that aspect words in Turkish reviews are usually the head words of the phrases in the sentences. Based on this observation and knowing that the head words in the dependency parse of a sentence appear close to the root of the parse tree, we decided to use this information in the aspect extraction models and integrated it into the position information. In this direction, instead of using the word positions in the sentence as positional encodings, we use the level numbers of the words in the dependency parse tree as positional encodings. As an example, the dependency parse tree of the sentence "\textit{Son olarak dün gittiğim her gittiğimde ayrı keyif aldığım güzel mekan.}" ("\textit{The beautiful place that I went yesterday lastly and that I enjoy every time I go.}") is shown in Fig. \ref{fig:tree} accompanied with the level indices. The root word is indexed with the length of the path between it and the most distant word (max-index) and the index of a word is set as (max-index - distance to the root). To index the words, we use a recursive algorithm starting with the root word in the dependency parser tree. In the example sentence, we see that the aspect term "mekan" ("place") is associated with level index 6 which differentiates it from non-aspect terms in the sentence that have lower index values.

To integrate the level information into the model, we modified the positional encodings as shown in Equations 3 and 4. The word position \textit{pos} is replaced with $level\_index(pos)$ which denotes the level index of the word in that position in the dependency parse tree. The dependency parse trees are obtained using the BOUN dependency parser as in Section \ref{sec:POS}.

\begin{equation}
	\label{eq:pe1}
	PE_{pos,2i} = \sin{\left(\frac{level\_index(pos)}{M^\frac{2i}{d}{}}\right)}
\end{equation}
\begin{equation}
	\label{eq:pe2}
	PE_{pos,2i+1} = \cos{\left(\frac{level\_index(pos)}{M^\frac{2i}{d}{}}\right)}
\end{equation}

\begin{figure}[th!]
	\centering
	\caption{\bf Dependency parse tree of the sentence "Son olarak dün gittiğim her gittiğimde ayrı keyif aldığım güzel mekan."}
	\includegraphics[]{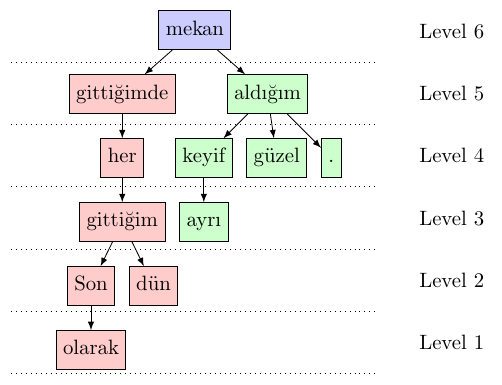}
	\label{fig:dtree}
\end{figure}

\subsection{BiLSTM and CRF Layers}

The word embedding, the POS tag embedding, and the tree positional encoding of each word in the sentence are concatenated and fed as input to the higher layers of the network. We build the network as a combination of BiLSTM and CRF models in order to effectively capture and label the aspects in the reviews. The BiLSTM layer processes the input words sequentially while capturing the contextual information and dependencies between the words. The LSTM cells maintain a hidden state and a cell state that evolve as new words are processed. The hidden state encodes contextual information from the previous words in the sequence, whereas the cell state holds information about long range dependencies.

The CRF layer takes the LSTM-generated features as input and assigns aspect labels to each word in the sentence by taking into account the relationships between neighboring words. To label the aspects, we use the BIO scheme \cite{semeval2016} which is a commonly used labeling scheme in sequence classification tasks. "B" marks the first term of an aspect, "I" marks the intermediate terms (terms except the first term) of the aspect, and "O" marks the non-aspect terms. The reason for using a CRF model on top of the LSTM model is that the CRF model considers the label transitions between the words and aims to find the optimal sequence of labels for the entire sentence. For each word in the sentence, the CRF model produces scores for the three aspect labels that reflect the compatibility of each label with the context. In addition, the transition probabilities from the aspect labels in a word position to the labels in the next position are kept in a transition matrix.

We employ an end-to-end training scheme in the sense that the error computed as the output of the CRF layer for each word is backpropagated through the entire network until the input embeddings and the parameters are updated. In the first model where random or Word2Vec embeddings are used as the initial input embeddings, the training scheme involves updating the initial embeddings. In the second model where the tokens are fed to the BERT model to produce the word embeddings, the training scheme indicates that the token embeddings as well as the parameters of BERT are updated. As a loss function, we use the negative log-likelihood loss which measures the difference between the predicted aspect label probabilities and true label probabilities. In this manner, the integrated model is trained to optimize a global objective function that considers the contributions of all the BERT, BiLSTM, and CRF models to the overall aspect extraction task.

\subsection{Viterbi Decoding}

During the test phase of the proposed models, we use the Viterbi decoding algorithm to obtain the aspect labels of the words in the sentence as output. The Viterbi algorithm \cite{viterbi1980} is a dynamic programming algorithm commonly used to find the most likely sequence of hidden states in a hidden Markov model or CRF model. In the context of sequence labeling tasks similar to aspect extraction, the algorithm determines the optimal sequence of labels for a given input sequence by considering the transition probabilities between the labels and the likelihood of the words. Beginning with the first word in the sentence, the algorithm iteratively selects the best aspect label for each word using these two probability distributions.

\subsection{Evaluation}

We evaluate the results of the models in terms of F1-values. The weighted-averaged F1-value is computed by averaging the F1 values of the classes (aspect labels) while considering the support of each class. The support of a class is the number of instances in the test set that belong to that class. The F1-value of a class is calculated using Equation 5, where TP, FP, and FN correspond to true positives, false positives, and false negatives, respectively.

\begin{equation}
	\label{eq:F1}
	F1 = \frac{TP}{TP + \frac{1}{2} (FP + FN)}
\end{equation}

\section{Experiments and Results}
\label{sec:experiments}

In this section, we explain the experiments conducted with the proposed models and give the performance of the models for Turkish aspect extraction. We consider the two models stated in Section \ref{sec:model} separately. We analyze their performance in detail with different input parameters and different initialization conditions of the parameters. We give the results on the standard train/test splits of the datasets in order to be able to compare the results with those in the literature. In addition, we also give the results obtained with k-fold cross-validation (k=5) to show the generalization performance of the models. 

Tables \ref{tab:lstm_turk} and \ref{tab:bert_turk} show the results on the SemEval 2016 Task 5 Restaurant Turkish Reviews dataset. Table \ref{tab:lstm_turk} depicts the result of the first model where random or Word2Vec embeddings are used for the word and POS tag embeddings. The first column denotes which inputs (words, POS tags, position information) are used (+) in the experiment. The second column shows how the word and POS tag embeddings are initialized, as random vectors or as Word2Vec vectors. The next column specifies the type of positional information used, where \textit{positional} stands for the positional encoding used in the original transformer model and TPE is the tree positional encoding proposed in this work. The last two columns are the results with the standard train/test splits of the datasets and with k-fold cross-validation.

We see that using random vectors or Word2Vec vectors to initialize the word and POS tag embeddings does not cause a consistent change in the results for both of the splits. Random initialization outperforms Word2Vec initialization in some cases while Word2Vec vectors show a better behavior in the others. This may be attributed to both types of the vectors being adapted to the task domain during the training phase of the aspect extraction models. Another observation is that the tree positional encodings show better performance than the original positional encodings. This result signals that using the distance of the words in a sentence from the head of the sentence provides useful information in identifying the aspects in the sentence.

Table \ref{tab:bert_turk} shows the results of the second model where the word embeddings are obtained with the BERT model. We observe a significant increase in the success rates compared to the model with random/Word2Vec word vectors. This result aligns with the literature in the sense that contextualized BERT embeddings outperform non-contextualized embeddings in downstream NLP tasks \cite{chauhan2023}. The table also shows that the positional encodings mostly contribute to the performance of the model with different types of POS tag embeddings. When we consider all the results on this dataset shown in Tables \ref{tab:lstm_turk} and \ref{tab:bert_turk}, we see that using BERT embeddings for the words and integrating tree positional encoding information to the network yield the best results with both the standard train/test split and k-fold cross-validation.

\begin{table}[h!]
		\centering
		\caption{BiLSTM-CRF model macro F1-values for Turkish SemEval 2016 Restaurant Reviews dataset.}
		\label{tab:lstm_turk}
		\setlength{\tabcolsep}{2pt}
		\begin{tabular}{|C{32pt}|C{28pt}|C{46pt}|C{63pt}|C{53pt}|C{76pt}|C{60pt}|}
			\addlinespace
			\hline 
			\multicolumn{3}{|c|}{\textbf{Input}} &
			\multirow{2}{*}{\splitl{\bf Word/POS}}
			&  \multirow{2}{*}{\splitl{\bf Position}}
			&  \multirow{2}{*}{\splitl{\textbf{Macro F-1} }}
			&  \multirow{2}{*}{\splitl{\textbf{Macro F-1}}} \\[+3pt]
			
			\begin{tabular}{c} \bf Word \end{tabular} &
			\begin{tabular}{c} \bf POS \\\bf Tag \end{tabular} &
			\begin{tabular}{c} \bf Position \end{tabular} &
			\begin{tabular}{l} \bf Embedding \end{tabular} & 
			\begin{tabular}{l} \bf Encoding \end{tabular} &
			\begin{tabular}{l} \bf Original Split \end{tabular} & 
			\begin{tabular}{l} \bf K-Fold\end{tabular}
			\\ \hline 
			\begin{tabular}{c} + \end{tabular} &
			\begin{tabular}{c} - \end{tabular} &
			\begin{tabular}{c} - \end{tabular} & 
			\begin{tabular}{c} Random \end{tabular} &
			\begin{tabular}{c} - \end{tabular} &
			\begin{tabular}{c} 46.39 \end{tabular} & 
			\begin{tabular}{c} 47.67 \end{tabular}
			\\ \hline
			\begin{tabular}{c} + \end{tabular} &
			\begin{tabular}{c} - \end{tabular} &
			\begin{tabular}{c} - \end{tabular} & 
			\begin{tabular}{c} Word2Vec \end{tabular} &
			\begin{tabular}{c} - \end{tabular} &
			\begin{tabular}{c} 50.63 \end{tabular} & 
			\begin{tabular}{c} 49.98 \end{tabular}
			\\ \hline
			\begin{tabular}{c} + \end{tabular} &
			\begin{tabular}{c} + \end{tabular} &
			\begin{tabular}{c} - \end{tabular} & 
			\begin{tabular}{c} Random \end{tabular} &
			\begin{tabular}{c} - \end{tabular} &
			\begin{tabular}{c} \textbf{54.65} \end{tabular} & 
			\begin{tabular}{c} 49.63 \end{tabular}
			\\ \hline
			\begin{tabular}{c} + \end{tabular} &
			\begin{tabular}{c} + \end{tabular} &
			\begin{tabular}{c} - \end{tabular} & 
			\begin{tabular}{c} Word2vec \end{tabular} &
			\begin{tabular}{c} - \end{tabular} &
			\begin{tabular}{c} 52.50 \end{tabular} & 
			\begin{tabular}{c} \textbf{52.13} \end{tabular}
			\\ \hline
			\begin{tabular}{c} + \end{tabular} &
			\begin{tabular}{c} - \end{tabular} & 
			\begin{tabular}{c} + \end{tabular} & 
			\begin{tabular}{c} Random \end{tabular} &
			\begin{tabular}{c} Positional \end{tabular} &
			\begin{tabular}{c} 47.95 \end{tabular} & 
			\begin{tabular}{c} 49.25 \end{tabular}
			\\ \hline
			\begin{tabular}{c} + \end{tabular} &
			\begin{tabular}{c} - \end{tabular} & 
			\begin{tabular}{c} + \end{tabular} & 
			\begin{tabular}{c} Word2vec \end{tabular} &
			\begin{tabular}{c} Positional \end{tabular} &
			\begin{tabular}{c} 50.00 \end{tabular} & 
			\begin{tabular}{c} 45.56 \end{tabular}
			\\ \hline
			\begin{tabular}{c} + \end{tabular} &
			\begin{tabular}{c} + \end{tabular} & 
			\begin{tabular}{c} + \end{tabular} & 
			\begin{tabular}{c} Random \end{tabular} &
			\begin{tabular}{c} Positional \end{tabular} &
			\begin{tabular}{c} 52.20 \end{tabular} & 
			\begin{tabular}{c} 47.39 \end{tabular}
			\\ \hline
			\begin{tabular}{c} + \end{tabular} &
			\begin{tabular}{c} + \end{tabular} & 
			\begin{tabular}{c} + \end{tabular} & 
			\begin{tabular}{c} Word2vec \end{tabular} &
			\begin{tabular}{c} Positional \end{tabular} &
			\begin{tabular}{c} 50.60 \end{tabular} & 
			\begin{tabular}{c} 42.03 \end{tabular}
			\\ \hline
			\begin{tabular}{c} + \end{tabular} &
			\begin{tabular}{c} - \end{tabular} & 
			\begin{tabular}{c} + \end{tabular} & 
			\begin{tabular}{c} Random \end{tabular} &
			\begin{tabular}{c} TPE \end{tabular} &
			\begin{tabular}{c} 51.08 \end{tabular} & 
			\begin{tabular}{c} 49.26 \end{tabular}
			\\ \hline
			\begin{tabular}{c} + \end{tabular} &
			\begin{tabular}{c} - \end{tabular} & 
			\begin{tabular}{c} + \end{tabular} & 
			\begin{tabular}{c} Word2vec \end{tabular} &
			\begin{tabular}{c} TPE \end{tabular} &
			\begin{tabular}{c} 52.83 \end{tabular} & 
			\begin{tabular}{c} 46.66 \end{tabular}
			\\ \hline
			\begin{tabular}{c} + \end{tabular} &
			\begin{tabular}{c} + \end{tabular} & 
			\begin{tabular}{c} + \end{tabular} & 
			\begin{tabular}{c} Random \end{tabular} &
			\begin{tabular}{c} TPE \end{tabular} &
			\begin{tabular}{c} 52.91 \end{tabular} & 
			\begin{tabular}{c} 51.41 \end{tabular}
			\\ \hline
			\begin{tabular}{c} + \end{tabular} &
			\begin{tabular}{c} + \end{tabular} & 
			\begin{tabular}{c} + \end{tabular} & 
			\begin{tabular}{c} Word2vec \end{tabular} &
			\begin{tabular}{c} TPE \end{tabular} &
			\begin{tabular}{c} 53.92 \end{tabular} & 
			\begin{tabular}{c} 50.40 \end{tabular}
			\\ \hline 
		\end{tabular}
\end{table}

\begin{table}[h!]
	\centering
		\caption{BERT-BiLSTM-CRF model macro F1-values for Turkish SemEval 2016 Restaurant Reviews dataset.}
		\label{tab:bert_turk}
		\setlength{\tabcolsep}{2pt}
		\begin{tabular}{|C{28pt}|C{46pt}|C{62pt}|C{52pt}|C{110pt}|C{60pt}|}
			\hline 
			\multicolumn{2}{|c|}{\textbf{Input}} &
			\multirow{2}{*}{\splitl{\bf POS}} &
			\multirow{2}{*}{\splitl{\bf Position}} &
			\multirow{2}{*}{\splitl{\textbf{Macro F-1 for Orig.} }} &
			\multirow{2}{*}{\splitl{\textbf{Macro F-1}}} \\[+3pt]
			
			\begin{tabular}{c} \bf POS \\\bf Tag \end{tabular} &
			\begin{tabular}{c} \bf Position \end{tabular} &
			\begin{tabular}{l} \bf Embedding \end{tabular} & 
			\begin{tabular}{l} \bf Encoding \end{tabular} &
			\begin{tabular}{l} \bf Train-Test Split \end{tabular} & 
			\begin{tabular}{l} \bf for k-Fold\end{tabular}
			\\ \hline
			\begin{tabular}{l} - \end{tabular} & 
			\begin{tabular}{l} - \end{tabular} &
			\begin{tabular}{l} - \end{tabular} & 
			\begin{tabular}{l} - \end{tabular} &
			\begin{tabular}{c} 74.02 \end{tabular} & 
			\begin{tabular}{c} 66.74 \end{tabular}
			\\ \hline
			\begin{tabular}{l} + \end{tabular} & 
			\begin{tabular}{l} - \end{tabular} &
			\begin{tabular}{c} Random \end{tabular} &
			\begin{tabular}{l} - \end{tabular} & 
			\begin{tabular}{c} 74.13 \end{tabular} & 
			\begin{tabular}{c} 67.53 \end{tabular}
			\\ \hline
			\begin{tabular}{l} + \end{tabular} & 
			\begin{tabular}{l} - \end{tabular} &
			\begin{tabular}{c} Word2vec \end{tabular} &
			\begin{tabular}{l} - \end{tabular} & 
			\begin{tabular}{c} 74.50 \end{tabular} & 
			\begin{tabular}{c} 68.36 \end{tabular}
			\\ \hline
			\begin{tabular}{l} - \end{tabular} & 
			\begin{tabular}{l} + \end{tabular} &
			\begin{tabular}{l} - \end{tabular} & 
			\begin{tabular}{l} Positional \end{tabular} &
			\begin{tabular}{c} 73.09 \end{tabular} & 
			\begin{tabular}{c} 67.70 \end{tabular}
			\\ \hline
			\begin{tabular}{l} + \end{tabular} & 
			\begin{tabular}{l} + \end{tabular} &
			\begin{tabular}{l} Random \end{tabular} & 
			\begin{tabular}{c} Positional \end{tabular} &
			\begin{tabular}{c} 73.17 \end{tabular} & 
			\begin{tabular}{c} 67.21 \end{tabular}
			\\ \hline
			\begin{tabular}{l} + \end{tabular} & 
			\begin{tabular}{l} + \end{tabular} &
			\begin{tabular}{l} Word2vec \end{tabular} & 
			\begin{tabular}{c} Positional \end{tabular} &
			\begin{tabular}{c} 74.75 \end{tabular} & 
			\begin{tabular}{c} 67.66 \end{tabular}
			\\ \hline
			\begin{tabular}{l} - \end{tabular} & 
			\begin{tabular}{l} + \end{tabular} &
			\begin{tabular}{l} - \end{tabular} & 
			\begin{tabular}{l} TPE \end{tabular} &
			\begin{tabular}{c} 68.89 \end{tabular} & 
			\begin{tabular}{c} 67.76 \end{tabular}
			\\ \hline
			\begin{tabular}{l} + \end{tabular} & 
			\begin{tabular}{l} + \end{tabular} &
			\begin{tabular}{l} Random \end{tabular} & 
			\begin{tabular}{c} TPE \end{tabular} &
			\begin{tabular}{c} 74.88 \end{tabular} & 
			\begin{tabular}{c} 68.34 \end{tabular}
			\\ \hline
			\begin{tabular}{l} + \end{tabular} & 
			\begin{tabular}{l} + \end{tabular} &
			\begin{tabular}{l} Word2vec \end{tabular} & 
			\begin{tabular}{c} TPE \end{tabular} &
			\begin{tabular}{c} \textbf{75.74} \end{tabular} & 
			\begin{tabular}{c} \textbf{68.73} \end{tabular}
			\\ \hline
		\end{tabular}
\end{table}

Tables \ref{tab:lstm_trans} and \ref{tab:bert_trans} show the results on the machine-translated dataset obtained from the SemEval 2016 Task 5 Restaurant English Reviews dataset. We see a similar behavior as in the other dataset. Using random vectors or Word2Vec vectors for word and POS tag embeddings does not make a significant difference in the results. Including POS information in the models consistently increases the performance compared to using only the word information. In the BERT-based model, tree positional encodings contribute to the performance with different types of POS tag embeddings and also when POS tag embeddings are not used. As in the previous dataset, the best results in this dataset are obtained with contextualized BERT embeddings, and using tree positional encoding yields the best performance.

\begin{table}[h!]
	\centering
		\caption{BiLSTM-CRF model macro F1-values for machine-translated English SemEval 2016 Restaurant Reviews dataset.}
		\label{tab:lstm_trans}
		\begin{tabular}{|C{32pt}|C{28pt}|C{48pt}|C{64pt}|C{52pt}|C{80pt}|C{60pt}|}
			\addlinespace
			\hline 
			\multicolumn{3}{|c|}{\textbf{Input}} &
			\multirow{2}{*}{\splitl{\bf Word/POS}}
			&  \multirow{2}{*}{\splitl{\bf Position}}
			&  \multirow{2}{*}{\splitl{\textbf{Macro F-1} }}
			&  \multirow{2}{*}{\splitl{\textbf{Macro F-1}}} \\[+3pt]
			
			\begin{tabular}{c} \bf Word \end{tabular} &
			\begin{tabular}{c} \bf POS \\\bf Tag \end{tabular} &
			\begin{tabular}{c} \bf Position \end{tabular} &
			\begin{tabular}{l} \bf Embedding \end{tabular} & 
			\begin{tabular}{l} \bf Encoding \end{tabular} &
			\begin{tabular}{l} \bf Original Split \end{tabular} & 
			\begin{tabular}{l} \bf K-Fold\end{tabular}
			\\ \hline 
			\begin{tabular}{c} + \end{tabular} &
			\begin{tabular}{c} - \end{tabular} &
			\begin{tabular}{l} - \end{tabular} &
			\begin{tabular}{l} Random \end{tabular} &
			\begin{tabular}{l} - \end{tabular} &
			\begin{tabular}{c} 55.96 \end{tabular} & 
			\begin{tabular}{c} 59.66 \end{tabular}
			\\ \hline
			\begin{tabular}{c} + \end{tabular} &
			\begin{tabular}{c} - \end{tabular} &
			\begin{tabular}{l} - \end{tabular} &
			\begin{tabular}{l} Word2Vec \end{tabular} &
			\begin{tabular}{l} - \end{tabular} &
			\begin{tabular}{c} 61.33 \end{tabular} & 
			\begin{tabular}{c} 61.69 \end{tabular}
			\\ \hline
			\begin{tabular}{c} + \end{tabular} &
			\begin{tabular}{c} + \end{tabular} &
			\begin{tabular}{l} - \end{tabular} &
			\begin{tabular}{l} Random \end{tabular} &
			\begin{tabular}{l} - \end{tabular} &
			\begin{tabular}{c} 60.62 \end{tabular} & 
			\begin{tabular}{c} 61.05 \end{tabular}
			\\ \hline
			\begin{tabular}{c} + \end{tabular} &
			\begin{tabular}{c} + \end{tabular} &
			\begin{tabular}{l} - \end{tabular} &
			\begin{tabular}{l} Word2vec \end{tabular} &
			\begin{tabular}{l} - \end{tabular} &
			\begin{tabular}{c} 61.89 \end{tabular} & 
			\begin{tabular}{c} \textbf{63.89} \end{tabular}
			\\ \hline
			\begin{tabular}{c} + \end{tabular} &
			\begin{tabular}{c} - \end{tabular} &
			\begin{tabular}{l} + \end{tabular} &
			\begin{tabular}{l} Random \end{tabular} & 
			\begin{tabular}{l} Positional \end{tabular} &
			\begin{tabular}{c} 60.76 \end{tabular} & 
			\begin{tabular}{c} 38.01 \end{tabular}
			\\ \hline
			\begin{tabular}{c} + \end{tabular} &
			\begin{tabular}{c} - \end{tabular} &
			\begin{tabular}{l} + \end{tabular} &
			\begin{tabular}{l} Word2vec \end{tabular} & 
			\begin{tabular}{l} Positional \end{tabular} &
			\begin{tabular}{c} 47.34 \end{tabular} & 
			\begin{tabular}{c} 30.22 \end{tabular}
			\\ \hline
			\begin{tabular}{c} + \end{tabular} &
			\begin{tabular}{c} + \end{tabular} &
			\begin{tabular}{l} + \end{tabular} &
			\begin{tabular}{l} Random \end{tabular} & 
			\begin{tabular}{l} Positional \end{tabular} &
			\begin{tabular}{c} 47.17 \end{tabular} & 
			\begin{tabular}{c} 34.90 \end{tabular}
			\\ \hline
			\begin{tabular}{c} + \end{tabular} &
			\begin{tabular}{c} + \end{tabular} &
			\begin{tabular}{l} + \end{tabular} &
			\begin{tabular}{l} Word2vec \end{tabular} & 
			\begin{tabular}{l} Positional \end{tabular} &
			\begin{tabular}{c} 60.62 \end{tabular} & 
			\begin{tabular}{c} 43.65 \end{tabular}
			\\ \hline
			\begin{tabular}{c} + \end{tabular} &
			\begin{tabular}{c} - \end{tabular} &
			\begin{tabular}{l} + \end{tabular} &
			\begin{tabular}{l} Random \end{tabular} & 
			\begin{tabular}{l} TPE \end{tabular} &
			\begin{tabular}{c} 59.87 \end{tabular} & 
			\begin{tabular}{c} 59.79 \end{tabular}
			\\ \hline
			\begin{tabular}{c} + \end{tabular} &
			\begin{tabular}{c} - \end{tabular} &
			\begin{tabular}{l} + \end{tabular} &
			\begin{tabular}{l} Word2vec \end{tabular} & 
			\begin{tabular}{l} TPE \end{tabular} &
			\begin{tabular}{c} 58.88 \end{tabular} & 
			\begin{tabular}{c} 57.61 \end{tabular}
			\\ \hline
			\begin{tabular}{c} + \end{tabular} &
			\begin{tabular}{c} + \end{tabular} &
			\begin{tabular}{l} + \end{tabular} &
			\begin{tabular}{l} Random \end{tabular} & 
			\begin{tabular}{l} TPE \end{tabular} &
			\begin{tabular}{c} \textbf{62.48} \end{tabular} & 
			\begin{tabular}{c} 62.12 \end{tabular}
			\\ \hline
			\begin{tabular}{c} + \end{tabular} &
			\begin{tabular}{c} + \end{tabular} &
			\begin{tabular}{l} + \end{tabular} &
			\begin{tabular}{l} Word2vec \end{tabular} & 
			\begin{tabular}{l} TPE \end{tabular} &
			\begin{tabular}{c} 59.54 \end{tabular} & 
			\begin{tabular}{c} 61.12 \end{tabular}
			\\ \hline 
		\end{tabular}
\end{table}

\begin{table}[h!]
		\centering
		\caption{BERT-BiLSTM-CRF model macro F1-values for machine-translated English SemEval 2016 Restaurant Reviews dataset.}
		\label{tab:bert_trans}
		\begin{tabular}{|C{28pt}|C{47pt}|C{64pt}|C{54pt}|C{115pt}|C{60pt}|}
			\hline 
			\multicolumn{2}{|c|}{\textbf{Input}} &
			\multirow{2}{*}{\splitl{\bf POS}} &
			\multirow{2}{*}{\splitl{\bf Position}} &
			\multirow{2}{*}{\splitl{\textbf{Macro F-1 for Orig.} }} &
			\multirow{2}{*}{\splitl{\textbf{Macro F-1}}} \\[+3pt]
			
			\begin{tabular}{c} \bf POS \\\bf Tag \end{tabular} &
			\begin{tabular}{c} \bf Position \end{tabular} &
			\begin{tabular}{l} \bf Embedding \end{tabular} & 
			\begin{tabular}{l} \bf Encoding \end{tabular} &
			\begin{tabular}{l} \bf Train-Test Split \end{tabular} & 
			\begin{tabular}{l} \bf for k-Fold\end{tabular}
			\\ \hline
			\begin{tabular}{l} - \end{tabular} &
			\begin{tabular}{l} - \end{tabular} &
			\begin{tabular}{l} - \end{tabular} &
			\begin{tabular}{l} - \end{tabular} &
			\begin{tabular}{c} 70.04 \end{tabular} & 
			\begin{tabular}{c} 65.96 \end{tabular}
			\\ \hline
			\begin{tabular}{l} + \end{tabular} &
			\begin{tabular}{l} - \end{tabular} &
			\begin{tabular}{l} Random \end{tabular} &
			\begin{tabular}{c} - \end{tabular} &
			\begin{tabular}{c} 71.89 \end{tabular} & 
			\begin{tabular}{c} 72.59 \end{tabular}
			\\ \hline
			\begin{tabular}{l} + \end{tabular} &
			\begin{tabular}{l} - \end{tabular} &
			\begin{tabular}{l} Word2vec \end{tabular} &
			\begin{tabular}{c} - \end{tabular} &
			\begin{tabular}{c} 71.29 \end{tabular} & 
			\begin{tabular}{c} 73.07 \end{tabular}
			\\ \hline
			\begin{tabular}{l} - \end{tabular} & 
			\begin{tabular}{l} + \end{tabular} &
			\begin{tabular}{l} - \end{tabular} & 
			\begin{tabular}{l} Positional \end{tabular} &
			\begin{tabular}{c} 71.15 \end{tabular} & 
			\begin{tabular}{c} 73.31 \end{tabular}
			\\ \hline
			\begin{tabular}{l} + \end{tabular} & 
			\begin{tabular}{l} + \end{tabular} &
			\begin{tabular}{l} Random \end{tabular} & 
			\begin{tabular}{c} Positional \end{tabular} &
			\begin{tabular}{c} 68.95 \end{tabular} & 
			\begin{tabular}{c} 72.61 \end{tabular}
			\\ \hline
			\begin{tabular}{l} + \end{tabular} & 
			\begin{tabular}{l} + \end{tabular} &
			\begin{tabular}{l} Word2vec \end{tabular} & 
			\begin{tabular}{c} Positional \end{tabular} &
			\begin{tabular}{c} 68.82 \end{tabular} & 
			\begin{tabular}{c} 72.90 \end{tabular}
			\\ \hline
			\begin{tabular}{l} - \end{tabular} &
			\begin{tabular}{l} + \end{tabular} &
			\begin{tabular}{l} - \end{tabular} &
			\begin{tabular}{l} TPE \end{tabular} &
			\begin{tabular}{c} 70.69 \end{tabular} & 
			\begin{tabular}{c} 72.23 \end{tabular}
			\\ \hline
			\begin{tabular}{l} + \end{tabular} &
			\begin{tabular}{l} + \end{tabular} &
			\begin{tabular}{l} Random \end{tabular} &
			\begin{tabular}{c} TPE \end{tabular} &
			\begin{tabular}{c} 71.07 \end{tabular} & 
			\begin{tabular}{c} 73.26 \end{tabular}
			\\ \hline
			\begin{tabular}{l} + \end{tabular} &
			\begin{tabular}{l} + \end{tabular} &
			\begin{tabular}{l} Word2vec \end{tabular} &
			\begin{tabular}{c} TPE \end{tabular} &
			\begin{tabular}{c} \textbf{72.38} \end{tabular} & 
			\begin{tabular}{c} \textbf{73.36} \end{tabular}
			\\ \hline 
		\end{tabular}
\end{table}

The best results we obtained on the standard train/test splits for the SemEval 2016 Task 5 Restaurant Turkish Reviews dataset and the machine-translated SemEval 2016 Task 5 Restaurant English Reviews dataset are, respectively, 75.74 and 72.38 F1 values. We share the results of the SemEval 2016 Task 5 competition for Turkish and English in Tables \ref{tab:semeval-turkish} and \ref{tab:semeval-english}, respectively, and compare with the results obtained in this work. Table \ref{tab:semeval-turkish} shows that the result of the proposed models used in this work outperforms the shared task results with a large margin on the original Turkish dataset. Table \ref{tab:semeval-english} lists the shared task results on the English dataset and our result on the machine-translated Turkish dataset. Although the datasets are different, we show the results in the same table to make a comparison. For the translated Turkish dataset, the result we obtained slightly falls behind the shared task results. This can be attributed to the translated nature of the dataset. The translations are performed in a controlled manner and post-processing steps are applied, however the automatic translation process brings some noise into the texts. Although we used translated reviews in the experiments for this dataset, the results are close to the top scores obtained in the shared task.

\begin{table}[h!]
	\centering
	\caption{Comparison of models for Turkish SemEval 2016 Restaurant Reviews dataset.}
	\label{tab:semeval-turkish}
	\begin{tabular}{|p{90pt}|p{60pt}|}
		\hline 
		\textbf{Model} 
		& \textbf{F1-value}
		\\ \hline
		\begin{tabular}{l} \bf Our Model \end{tabular} & 
		\begin{tabular}{c} \bf 75.74 \end{tabular}
		\\ \hline
		\begin{tabular}{l} UFAL\cite{ufal2016} \end{tabular} & 
		\begin{tabular}{c} \bf 61.03 \end{tabular}
		\\ \hline
		\begin{tabular}{c} Wang et al.\cite{Wang2020} \end{tabular} &
		\begin{tabular}{c} 59.30 \end{tabular}
		\\ \hline
		\begin{tabular}{c} IIT-T\cite{iit-tuda2016} \end{tabular} &
		\begin{tabular}{c} 56.63 \end{tabular}
		\\ \hline
		\begin{tabular}{c} INSIG\cite{insig2016} \end{tabular} &
		\begin{tabular}{c} 49.12 \end{tabular}
		\\ \hline
	\end{tabular}
\end{table}

\begin{table}[h!]
	\centering
	\caption{Comparison of models for English SemEval 2016 Restaurant Reviews dataset.}
	\label{tab:semeval-english}
	\setlength{\tabcolsep}{3pt}
	\begin{tabular}{|p{70pt}|p{50pt}|p{95pt}|}
		\hline 
		\textbf{Model} 
		& \textbf{F1-value}
		& \textbf{Language}
		\\ \hline
		\begin{tabular}{l} Wang et al. \cite{Wang2021} \end{tabular} & 
		\begin{tabular}{c} \bf 81.30 \end{tabular} &
		\begin{tabular}{l} English \end{tabular}
		\\ \hline
		\begin{tabular}{l} Wei et al. \cite{Wei2020} \end{tabular} & 
		\begin{tabular}{c} 78.00 \end{tabular} &
		\begin{tabular}{l} English \end{tabular}
		\\ \hline
		\begin{tabular}{l} Xu et al. \cite{Xu2019} \end{tabular} & 
		\begin{tabular}{c} 77.70 \end{tabular} &
		\begin{tabular}{l} English \end{tabular}
		\\ \hline
		\begin{tabular}{c} Xu et al.\cite{Xu2018} \end{tabular} &
		\begin{tabular}{c} 75.40 \end{tabular} &
		\begin{tabular}{l} English \end{tabular}
		\\ \hline
		\begin{tabular}{c} Wang et al.\cite{Wang2020} \end{tabular} &
		\begin{tabular}{c} 72.80 \end{tabular} &
		\begin{tabular}{l} English \end{tabular}
		\\ \hline
		\begin{tabular}{c} \bf Our Model \end{tabular} &
		\begin{tabular}{c} \bf 72.38 \end{tabular} &
		\begin{tabular}{l} Turkish (Translated) \end{tabular}
		\\ \hline 
	\end{tabular}
\end{table}

\section{Conclusion}
\label{sec:conclusion}
In this work, we focused on the aspect extraction component of aspect-based sentiment analysis and developed novel aspect extraction models. The models integrate different embedding schemes for words and POS tags and different learning algorithms. By employing contextualized representations, the BERT embeddings contributed to the model's ability to distinguish complex relationships between words in a given context. The BiLSTM model addressed the sequential structure in the aspect terms and allowed the model to grasp long-term dependencies. Including CRF as the final layer increased the precision of the model by imposing constraints between adjacent predictions. Finally, tree positional encoding provided information about the positions of aspects in sentences.

We also built a new Turkish aspect extraction dataset by machine translating the Semeval 2016 Task 5 Restaurant English Reviews dataset into Turkish. After the dataset is translated, several post-processing operations were performed to correct the translation errors and to maintain the consistency of the dataset. The aspect positions were relabelled manually since the positions of the corresponding words in sentences in the two languages are different.

The experiments were conducted on the SemEval 2016 Turkish Restaurant Reviews dataset and the machine-translated Turkish dataset. We used both the original train/test splits of the datasets to compare the results with those in the literature and also k-fold cross-validation to generalize the results. We obtained around 72-75\% F1 values on the two datasets. The performance on the original Turkish dataset outperformed the success rates of the models in the shared task with a large margin. Although the machine-translated dataset is a new dataset, we also compared the performance of the proposed models on this dataset with the performance of the models in the shared task on the English version of the dataset. The proposed models fall behind the success rates of the other models. This may be regarded as an expected result since the translated dataset includes noise. We also observed that using tree positional encoding mostly outperforms the classical positional encoding. This result signals that the positions of words in the dependency hierarchy of the sentences give an important clue as being an aspect or not.

Recently, large language models have begun to be used in several NLP domains and they provide performance comparable to other classical and deep-learning-based approaches. As future work, we aim to employ large language models with different prompting methods and observe their performance in the aspect extraction task. We also plan to integrate the aspect extraction models proposed in this work to the existing sentiment analysis models in Turkish to arrive at an end-to-end aspect-based sentiment analysis framework.

\bibliographystyle{unsrt}  


\end{document}